\pdfoutput=1
\documentclass[11pt]{article}

\usepackage[]{acl}

\usepackage{times}
\usepackage{latexsym}

\usepackage[T1]{fontenc}
\usepackage[utf8]{inputenc}

\usepackage{microtype}

\usepackage{inconsolata}

\usepackage{hyperref}
\usepackage{url}

\usepackage[utf8]{inputenc} % allow utf-8 input
\usepackage[T1]{fontenc}    % use 8-bit T1 fonts
\usepackage{hyperref}       % hyperlinks
\usepackage{url}            % simple URL typesetting
\usepackage{booktabs}       % professional-quality tables
\usepackage{amsfonts}       % blackboard math symbols
\usepackage{nicefrac}       % compact symbols for 1/2, etc.
\usepackage{microtype}      % microtypography
\usepackage{xcolor}         % colors

\usepackage{graphicx}
\usepackage{mathtools}
\usepackage{wrapfig}
\usepackage{booktabs}
\usepackage{makecell}
\usepackage{pifont}

\usepackage[ruled,vlined]{algorithm2e}

\SetCommentSty{mycommfont}
\definecolor{mypink}{RGB}{255, 143, 171}
\definecolor{mygreen}{RGB}{86, 171, 145}
\usepackage{multirow}
\usepackage{enumitem}
\usepackage{subfiles}

\title{Expedited Training of Visual Conditioned Language Generation via Redundancy Reduction}

\author{Yiren Jian\thanks{This work was done during an internship at ByteDance Inc.} \\
  Dartmouth College
  \\\And
  Tingkai Liu \\
  ByteDance Inc.
  \\\And
  Yunzhe Tao \\
  ByteDance Inc.
  \\\AND
  Chunhui Zhang \\
  Dartmouth College
  \\\And
  Soroush Vosoughi \\
  Dartmouth College
  \\\And
  Hongxia Yang \\
  ByteDance Inc.}

\begin{document}
\maketitle
\begin{abstract}
In this paper, we introduce $\text{EVL}_{\text{Gen}}$, a streamlined framework designed for the pre-training of visually conditioned language generation models with high computational demands, utilizing frozen pre-trained large language models (LLMs). The conventional approach in vision-language pre-training (VLP) typically involves a two-stage optimization process: an initial resource-intensive phase dedicated to general-purpose vision-language representation learning, focused on extracting and consolidating relevant visual features. This is followed by a subsequent phase that emphasizes end-to-end alignment between visual and linguistic modalities. Our novel one-stage, single-loss framework bypasses the computationally demanding first training stage by gradually merging similar visual tokens during training, while avoiding model collapse caused by single-stage training of BLIP-2 type models. The gradual merging process effectively condenses visual information while preserving semantic richness, resulting in rapid convergence without compromising performance. Our experimental findings demonstrate that our approach accelerates the training of vision-language models by a factor of 5 without a noticeable impact on overall performance. Furthermore, we illustrate that our models significantly narrow the performance gap to current vision-language models using only 1/10 of the data. Finally, we showcase how our image-text models can seamlessly adapt to video-conditioned language generation tasks through novel soft attentive temporal token contextualizing modules. Code is available at \url{https://github.com/yiren-jian/EVLGen}.
\end{abstract}

\section{Introduction}
The landscape of vision-language modeling has undergone significant transformations in recent years, with CLIP~\citep{radford2021learning} serving as a landmark development. It distinguished itself through unparalleled zero-shot classification capabilities and efficiency in image-text retrieval tasks. Successive models like ALBEF~\citep{li2021align}, X-VLM~\citep{zeng2022multi}, and VLMo~\citep{bao2022vlmo} further broadened the scope, addressing a myriad of tasks such as retrieval, visual entailment, and closed-set Visual Question Answering (VQA), among others.

Recently, the field has been enriched by the advent of generative models designed for complex image-to-language tasks. Notable contributions include CoCa~\citep{yu2022coca}, SimVLM~\citep{wang2022simvlm}, Frozen~\citep{tsimpoukelli2021multimodal}, and Flamingo~\cite{alayrac2022flamingo}, targeting tasks like image and video captioning and open-set VQA. 
These models all rely on billion-scale datasets for training from scratch to bridge the substantial modality gap between vision and language.

As a result, the resource-intensive requirements (i.e., thousands of TPUs) of these training-from-scratch Vision-Language Models (VLMs) led to the conceptualization of BLIP-2~\citep{li2023blip}:
this model alleviates computational costs (e.g., only requiring $16\times$ fewer GPUs) by integrating existing well-pretrained vision encoders (ViT) with language decoders (LLM), and then tuning their joint operation. A central innovation in aligning vision and language modules in BLIP-2 is \textit{Q-former}, a multimodal connector equipped with learnable queries for enhancing cross-attention mechanisms. This architectural choice, however, prevents the full model from end-to-end training and therefore \textit{still} demands an additional pre-training regimen for Q-former, referred to as \textit{BLIP-2's Stage 1}. The stage involves three learning objectives—image-text contrastive, image-text matching, and language generation—and necessitates multiple forward passes for facilitating the Q-former's optimization.

\begin{figure*}[!t]
\centering
\includegraphics[width=0.8\textwidth]{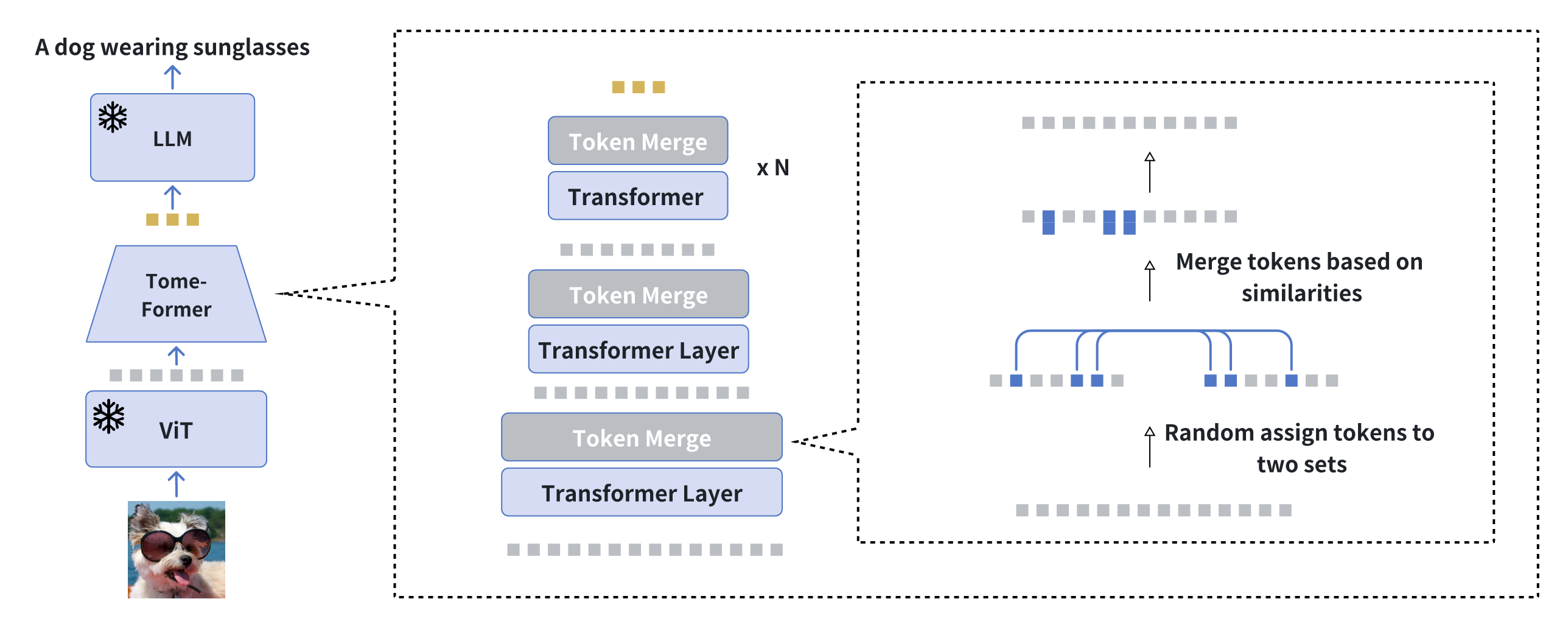}
\caption{Overview of our $\text{EVL}_{\text{Gen}}$. $\text{EVL}_{\text{Gen}}$ employs a streamlined, single-stage training mechanism with a unified loss. Here, visual tokens (in grey) are progressively aggregated based on their inherent similarities at each layer of the TomeFormer architecture. The final set of merged tokens (in orange) serves as semantically rich but computationally efficient soft prompts, guiding the LLM to generate a corresponding caption for the input image.}
\label{fig:overview}
\end{figure*}

Despite its efficiency gains over CoCa, BLIP-2's training still imposes considerable computational costs. This poses challenges for research environments with limited computational resources, such as university labs. Our experiments indicate that the Stage-1 training of BLIP-2 took approximately eight days on eight A100-80G GPUs (See Appendix~\ref{appendix:blip2-training-details} for training configurations). This computational burden has consequently restricted research to using the pre-trained Q-former, hindering the exploration of alternative ViTs in VLMs. This limitation is evident in subsequent works such as InstructBLIP~\citep{instructblip}, VideoChat~\citep{2023videochat}, Video-LLaMA~\citep{damonlpsg2023videollama}, X-LLM~\citep{chen2023xllm}.

The prospect of reducing BLIP-2's computational cost through end-to-end, single-stage training is compelling. Such an approach would remove the complexities associated with resource allocation and hyper-parameter tuning inherent in multi-stage training. Yet, direct end-to-end training with BLIP-2 poses substantial challenges, corroborated by both original findings from BLIP-2 and our own empirical analyses. We hypothesize that these challenges emanate from the intrinsic design of the Q-former. Specifically, the inclusion of randomly initialized learnable queries and cross-attention mechanisms complicates the optimization landscape, especially when the aim is to minimize the representational disparity between visual and linguistic modalities.

In this paper, we propose a token merging Transformer (TomeFormer) as an efficient vision-language connector. TomeFormer employs a systematic token-merging~\citep{bolya2022tome} strategy that is both intuitive and effective. By connecting a pre-trained ViT as the visual encoder and a frozen LLM as the language decoder, we introduce a new VLM ``\textbf{E}xpedited \textbf{V}isual \textbf{L}anguage \textbf{Gen}eration model'' ($\text{EVL}_{\text{Gen}}$), facilitates a streamlined, single-stage training process. It requires only a singular learning objective and a single forward pass per optimization step. This stands in contrast to BLIP-2's multi-stage training, laden with multiple objectives and several forward passes.

Further, we introduce a \textit{soft attentive temporal} contextualization mechanism within the ViT for effective video-language modeling. This uncovers more shared semantic features across temporal frames, thereby improving the efficiency of the spatial token merging process. It eliminates the need for modality realignment, contrasting approaches such as the temporal Q-former~\citep{damonlpsg2023videollama}, or the addition of new learnable temporal queries~\citep{2023videochat}. Our strategy simplifies the optimization challenges tied to working with relatively smaller video-text datasets, compared to their image-text counterparts. Remarkably, we demonstrate that even without video pre-training, our temporal token contextualize approach can effectively train robust video-language models. This differs from recent work in video-language models~\cite{yan2022videotext, chen2023vast} that depend on pre-training models using vast million-scale video-text datasets.
In summary, our contributions are:
\begin{itemize}[leftmargin=*,noitemsep,nolistsep]
    \item For reducing vision redundancy within the vision language connector, we adopt Token Merging, initially designed to enhance ViT inference speed without training. Concurrently, we present a novel temporal token contextualization scheme for video modeling.

    \item Our proposed VLM featuring TomeFormer competes effectively with BLIP-2, while requiring just a fraction of the computational resources. Given the reliance on BLIP-2's pre-trained model in contemporary studies, our approach widens the exploratory scope for various ViTs.

    \item We introduce a straightforward spatial attentive temporal modeling technique that allows for the seamless adaptation of pre-trained image-text models to video tasks. This approach eliminates the need for complex modality re-alignment, a common requirement in alternative methods.
    
\end{itemize}

\section{Related Work}
\paragraph{Image-Language Models} CoCa~\citep{yu2022coca}, trained on billions of image-text pairs, represents a state-of-the-art approach in generative tasks like open VQA and visual captioning. To mitigate the computational demands of pre-training, BLIP-2~\citep{li2023blip} employs frozen pre-trained ViT and LLM components, focusing on training a specialized connector between visual and linguistic modalities called the Q-former. Due to the computationally intensive nature of training BLIP-2, subsequent models in visual instruction \citep{instructblip, zhu2023minigpt, 2023videochat} have predominantly utilized the pre-trained Q-former, which is aligned with the \texttt{eva-vit-g} model supplied by BLIP-2. Additional related works on image-language modeling are further discussed in Appendix~\ref{appendix:additional_ref}.

\paragraph{Video-Language Models} While many image-text models can be adapted for video-text tasks through simple feature pooling (e.g., VideoCoCa~\citep{yan2022videotext}), the field has seen specialized models that incorporate temporal dynamics. Building on the foundation of BLIP-2, Video-LLaMA~\citep{damonlpsg2023videollama} enhances its architecture by introducing additional temporal Q-former layers between the spatial Q-former and the LLM components of BLIP-2. Inspired by BLIP-2, most recent works such as VideoChat~\citep{2023videochat}, PandaGPT~\citep{su2023pandagpt}, Valley~\citep{luo2023valley}, and Video-ChatGPT~\citep{Maaz2023VideoChatGPT} leverage frozen LLMs in their video-language models.

\paragraph{Token Merging} Token Merging (ToMe)~\citep{bolya2022tome} aims to improve the inference speed of pre-trained ViTs without requiring re-training. At each Transformer layer, tokens are divided into two sets and subsequently merged based on similarity, effectively reducing the token count and thereby accelerating inference. This method maintains classification and generation quality.

In our work, we repurpose ToMe to condense the visual features used as language prompts in the LLM. We integrate a standard Transformer with ToMe capabilities, resulting in a model we term TomeFormer. This model serves as an effective connector between visual and language domains, preserving semantic richness while reducing token count. Importantly, this integration of ToMe does not introduce any additional parameters. Inspired by spatial ToMe, we introduce a novel soft temporal ToMe variant within the vision encoder, thereby adding temporal modeling capabilities to our image-text models. Additional related works on token redundancy are further discussed in Appendix~\ref{appendix:additional_ref}.

\begin{figure*}[!t]
\centering
\includegraphics[width=0.8\textwidth]{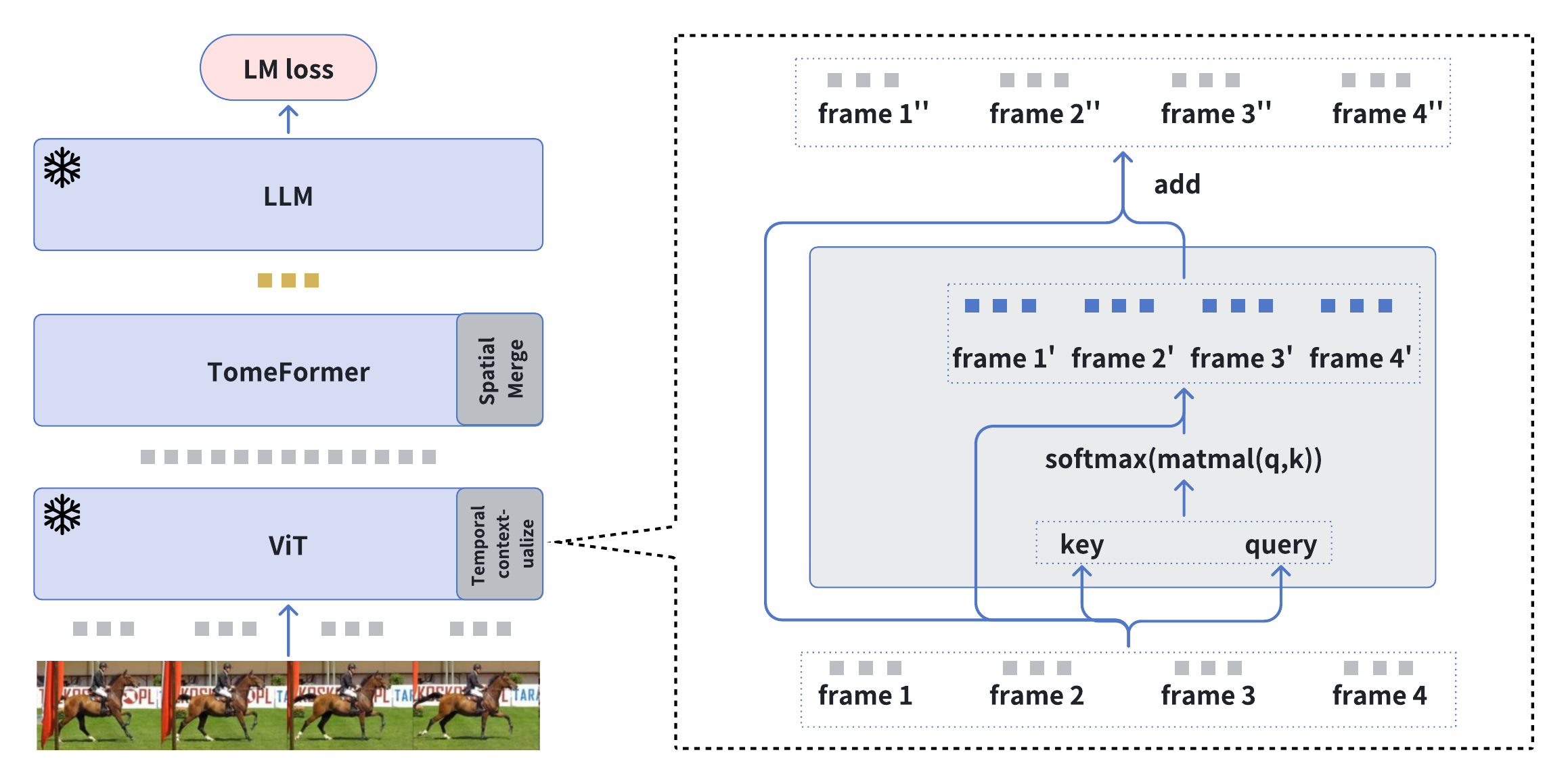}
\caption{Overview of $\text{EVL}_{\text{Gen}}$-Video: In addition to TomeFormer's spatial token merging capabilities, our design introduces Temporal Attentive Soft Token Contextualizing for nuanced temporal modeling. Each frame's output is calculated as a learnable weighted average of other frames in the video. This approach maintains the integration of pre-existing, well-trained image-text models. For instance, when the input consists of static videos with identical frames, $\text{EVL}_{\text{Gen}}$-Video operates as if it were an image-text model. Importantly, this architecture avoids the need for complex modality realignment, a requirement in alternative designs that insert a temporal Q-former between the visual encoder and the language model.
It also significantly enriches the shared semantic information distributed among these frame tokens, laying the groundwork for more efficient token merging in future spatial merging steps.
}
\label{fig:simvlg_video}
\end{figure*}

\section{Methods}
We begin by presenting our image-text model and then describe the adaptations made to this pre-trained model for video-related tasks.

\subsection{Preliminary}
We follow BLIP-2's efficient training paradigm, i.e., utilizing frozen but well-pretrained ViTs and LLMs while \emph{solely training a vision-to-language connector}. However, BLIP-2 still remains a challenge, since it necessitates an extra Stage-1 as a pre-training phase for the unstable Q-former (i.e., the vision-to-language connector), before the final end-to-end fine-tuning.

Our observation underscores the pivot role of BLIP-2 Stage-1 pre-training (which takes approximately 8 days on eight A100 GPUs): without it, the BLIP-2 model collapses, as evidenced in Table~\ref{table:vqa_mscoco_zeroshot}. To avoid this extra stage-1, we replace Q-former with a novel vision-to-language connector, which is designed to discover vision redundancy and then significantly accelerate visual-language alignment, often resulting in enhanced performance.

\subsection{$\text{EVL}_{\text{Gen}}$-Image}
We introduce $\text{EVL}_{\text{Gen}}$-Image (abbreviated as $\text{EVL}_{\text{Gen}}$, shown in Figure~\ref{fig:overview}), an optimized vision-language generative pre-training model. $\text{EVL}_{\text{Gen}}$ utilizes a ViT for visual encoding and an LLM for linguistic decoding. The key innovation is the incorporation of a standard Transformer, augmented with spatial Token Merging, to act as the connector between the visual and linguistic modalities.

Formally, our framework includes a vision encoder $E_{\text{vision}}$, which ingests an input image $I$ and encodes it into a fixed set of visual tokens: $[v_{1}, v_{2}, ... v_{L}] = E_{\text{vision}}(I)$. Here, $L$ denotes the number of image patches. Subsequently, we employ a Transformer equipped with token-merging modules (further technical details are provided in Appendix~\ref{appendix:tome_details}), termed as \textit{TomeFormer} ($T_{\text{v}\rightarrow\text{l}}$) as the vision-to-language connector. This module effectively compresses the token count: 
\begin{align}
    [v'_{1}, v'_{2}, ... v'_{L'}] = T_{\text{v}\rightarrow\text{l}}(f_{\text{proj}_{1}}([v_{1}, v_{2}, ... v_{L}])).
\end{align}
In this equation, $L'$ is considerably smaller than the initial token count $L$\footnote{We merge a fixed number of tokens at each layer of the TomeFormer. Finally, 256 visual tokens are reduced to 28 tokens. Ablation on merged tokens at each layer is studied in Section~\ref{sec:analysis}.}. The LLM decoder then employs these compressed tokens as soft prompts for text generation: 
\begin{align}
    \text{output} = D_{\text{LLM}}(f_{\text{proj}_{2}}([v'_{1}, v'_{2}, ... v'_{L'}])).
\end{align}
Projection functions $f_{\text{proj}_{1}}$ and $f_{\text{proj}_{2}}$ are used to ensure dimension compatibility. The training objective is to minimize the cross-entropy between the output and ground truth caption: 
\begin{align}
    \mathcal{L} = \text{CrossEntropyLoss}(\text{output}, \text{cap}_{\text{gt}}).
\end{align}
Three main advantages of using TomeFormer are:
\begin{itemize}[leftmargin=*,noitemsep,nolistsep]
    \item Efficient token reduction, facilitating the transformation of loosely structured visual data into a more concise yet informative representation.
    
    \item Computational efficiency, as the uncompressed ViT output consists of 256 tokens, plus a [CLS] token. Without compression, the subsequent vision-to-language connector would be computationally expensive in terms of both memory and processing power.
    
    \item Semantic richness of the compressed tokens. Unlike BLIP-2, which requires an extensive pre-training phase for Q-former, TomeFormer naturally merges semantically similar tokens. Our empirical evidence confirms that TomeFormer-equipped models train more efficiently compared to alternatives like BLIP-2.
\end{itemize}

\subsection{$\text{EVL}_{\text{Gen}}$-Video}

Although many image-text models can be adapted for video-text tasks with minor modifications, such adaptations either result in inadequate temporal modeling (as in VideoCoCa or InstructBLIP) or require re-alignment with substantial video-text pairs due to additional learnable Q-formers (as in VideoChat and Video-LLaMA).

In this paper, we propose a novel module called \textit{Temporal Attentive Soft Token Contextualizing} to enhance the ViT backbone with temporal modeling capabilities (depicted in Figure~\ref{fig:simvlg_video}). A key feature of temporal soft contextualizing is that it is equivalent to the identity operator when the input is static images instead of videos. Thus, our approach maintains the integration of pre-existing, well-trained image-text models, thus avoiding the additional need for modality realignment, a requirement in alternative designs that insert a temporal Q-former between the visual encoder and LLM.

Formally, let $v$ be a video feature tensor with dimensions $[B \times N \times L \times D]$, where $B$ is the batch size, $N$ is the number of frames, $L$ is the sequence length (i.e., the number of patches in a single video frame), and $D$ is the hidden dimension. Initially, we reshape $v$ into $[(B \times N) \times L \times D]$ which is subsequently fed into the self-attention layer of the ViT for \textit{spatial modeling} as:
\begin{align}
    v' = \text{self-attn}(v.\text{reshape}(B \times N, L, D)).
\end{align}
For \textit{temporal modeling}, $v'$ is reshaped to $[(B \times L), N, D]$. We then project this into key and query matrices $k$ and $q$ and compute $v''$ using our \textit{Temporal Attentive Soft Token Contextualizing} as follows: 
\begin{align}
    &k = W_{\text{key}}(v'.\text{reshape}(B \times L, N, D)), \\
    &q = W_{\text{query}}(v'.\text{reshape}(B \times L, N, D)), \\
    &v'' = v' + \text{softmax}(\text{matmal}(q,k)) \cdot v'.
\end{align}
The softmax operation models temporal weights and \textit{softly} fuses tokens among multiple frames. This is distinct from spatial token merging, which employs average pooling and reduces the token count. Here, the weighted average pooling is applied to multiple frames for contextualization. It preserves the original count of tokens while enhancing the shared semantic content that is spread across various frames. Therefore, it allows a higher rate of token merging in the subsequent spatial merging processes.

\begin{table*}[!ht]
        \centering	
        \small
        \begin{tabular}	{l  l  l  l  l|  c  c  c  c  c}
        \toprule
        \multirow{2}{*}{Models} &\multirow{2}{*}{\makecell[l]{\# pre-train \\ image-text}} & \multirow{2}{*}{\makecell[l]{\# trainable \\ params}} & \multirow{2}{*}{\makecell[l]{\# stage-1 \\ steps}} & \multirow{2}{*}{\makecell[l]{\# stage-2 \\ steps}} & VQAv2 & GQA  & OK-VQA & COCO & \multirow{2}{*}{\makecell[c]{Clock \\time}}\\
        & & & & & val & test-dev & test & val\\
        \midrule
        VL-T5 & 9.2M & 224M & - & - & 13.5 & 6.3 & 5.8 & - & -\\
        FewVLM & 9.2M & 740M & - & - & 47.7 & 29.3 & 16.5 & - & -\\
        Frozen & 3M & 40M & - & - & 29.6 & - & 5.9 & - & - \\
        VLKD & 3M & 406M & - & - & 42.6 & - & 13.3 & - & -\\
        \midrule
        BLIP-2 & 104M$^{\dagger}$ & 110M+$^{\ddagger}$ & - & 80k/250k$^{*}$ & \ding{55} & \ding{55} & \ding{55} & \ding{55} & \ding{55} \\
        BLIP-2 & 104M & 110M+ & 250k & 80k & 44.6 & 30.6 & 26.0 & 137.7 & 234 hrs\\
        $\text{EVL}_{\text{Gen}}$ & 104M  & 55M & - & 90k  & 45.9 & 30.6 & 25.8 & 134.0 & 47 hrs\\
        $\text{EVL}_{\text{Gen}}$ & 11M$^{\dagger}$  & 110M & - & 150k & 46.3 & 30.0 & 23.0 & 135.1 & 80 hrs\\
        $\text{EVL}_{\text{Gen}}$ & 104M & 110M & - & 150k & 46.9 & 30.8 & 24.8 & 137.0 & 80 hrs\\
        $\text{EVL}_{\text{Gen}}$ & 104M & 110M & - & 250k & \textbf{48.4} & \textbf{30.9} & \textbf{27.2} & \textbf{139.1} & 133 hrs\\
        \bottomrule	
        \end{tabular}
        \caption{Comparison of methods on zero-shot VQA and MSCOCO captioning (CIDEr) tasks without additional fine-tuning. Both BLIP-2 and $\text{EVL}_{\text{Gen}}$ use OPT-2.7b as the LLM decoder. $^{*}$: \emph{BLIP-2 without extensive stage-1 pre-training will collapse}. $^{\dagger}$: We were only able to download approximately 81\% of LAION-115M (110M) and 78\% of CCS-14M (11M) from the CapFilt dataset. $^{\ddagger}$: BLIP-2 incorporates an additional set of 32 learnable queries, each with a dimension of 768.}
	\label{table:vqa_mscoco_zeroshot}
\end{table*}

\section{Experiments}
Our experimental setup is as follows:

\begin{itemize}[leftmargin=*,noitemsep,nolistsep]

\item \textbf{Pre-training Data} Our model is pre-trained using the MSCOCO~\citep{lin2014microsoft} and CapFilt~\citep{li2022blip} datasets, which include BLIP's pseudo-labeled Conceptual Captioning~\citep{sharma2018conceptual}, SBU~\citep{ordonez2011im2text}, and LAION~\citep{schuhmannlaion} datasets—similar to the data sources utilized in BLIP-2. Note that we intentionally exclude the VG~\citep{krishna2017visual} dataset from our pre-training procedure, as it mainly consists of localized captions. 

\item \textbf{Models} To facilitate a direct and fair comparison with BLIP-2, we employ the same ViT, texttt{eva-vit-g}~\citep{fang2022eva}. For the language model decoders, we explore both \texttt{opt-2.7b}~\citep{zhang2022opt} and \texttt{vicuna-7b}~\citep{vicuna2023}. Our TomeFormer is initialized using \texttt{bert-base-uncased}, ensuring parameter count parity with BLIP-2's Q-former.

\item \textbf{Pre-training Details} Our pre-training setup closely mirrors the configurations of BLIP-2. We utilize a maximum learning rate of \(1e^{-4}\) and a minimum learning rate of \(1e^{-5}\). The learning rate follows a schedule that begins with a linear warm-up phase of 5000 steps starting from \(1e^{-6}\) and then transitions to a cosine decay schedule. Weight decay is set at 0.05. The training is conducted with a batch size of 1600, distributed over either \(8 \times\) A100-80G or \(32 \times\) V100-32G. 

\item \textbf{Downstream Tasks} $\text{EVL}_{\text{Gen}}$-Image is evaluated without additional fine-tuning on a variety of tasks, including MSCOCO captioning, VQAv2~\citep{goyal2017making}, GQA~\citep{hudson2019gqa}, and OK-VQA~\citep{okvqa}. For video tasks, $\text{EVL}_{\text{Gen}}$-Video is evaluated on fine-tuned MSR-VTT~\citep{xu2016msr} and MSVD~\citep{chen2011collecting} captioning. We use the standard train/val/test splits.
\end{itemize}

\subsection{Evaluation on Image-Text Benchmarks}
We conducted comparative evaluations between $\text{EVL}_{\text{Gen}}$ and BLIP-2 on multiple image-text benchmarks, including zero-shot VQAv2, GQA, OK-VQA, and MSCOCO captioning. It is essential to note that BLIP-2 demands an extensive Stage-1 pre-training phase involving 250,000 optimization steps. This phase incorporates three distinct loss functions and necessitates multiple forward passes through the model, a process crucial for BLIP-2 to prevent model divergence.

Table~\ref{table:vqa_mscoco_zeroshot} summarizes the results of our experiments. Our primary insights can be distilled into the following key points:
\begin{itemize}[leftmargin=*,noitemsep,nolistsep]
    \item Utilizing the same training set of 104M image-text pairs and an equal number of optimization steps (250K), $\text{EVL}_{\text{Gen}}$ consistently outperforms BLIP-2 across nearly all evaluated tasks.
    \item Remarkably, $\text{EVL}_{\text{Gen}}$ maintains competitive performance even when its training budget is trimmed to approximately one-third of BLIP-2's, specifically 150K optimization steps.
    \item Our experiments show that $\text{EVL}_{\text{Gen}}$ can produce satisfactory results with a significantly reduced training dataset of 11 million image-text pairs, while still undergoing 150K optimization steps.
    \item $\text{EVL}_{\text{Gen}}$ retains its efficacy even when the training budget is restricted to as few as 90K steps, showing the model's efficiency and robustness.
\end{itemize}

We further evaluate BLIP-2 and $\text{EVL}_{\text{Gen}}$ on zero-shot NoCaps and Flickr30K datasets. Shown in Table~\ref{table:image_caption}, $\text{EVL}_{\text{Gen}}$ consistently outperforms BLIP-2 in both datasets using different LLMs.

\paragraph{Training Time} In the Stage-1 pre-training phase, BLIP-2 requires considerable time, necessitating multiple forward passes to optimize three separate loss functions. We document the training durations for both BLIP-2 and $\text{EVL}_{\text{Gen}}$ when utilizing eight A100-80G GPUs in the last column of Table~\ref{table:vqa_mscoco_zeroshot}. 

Although BLIP-2 significantly reduces training time relative to predecessors like CoCa, it still mandates an extended training duration, approximately ten days (8 days for stage 1 and 2 days for stage 2). This extensive time commitment limits the feasibility of researchers to investigate various ViT configurations. Most subsequent works based on BLIP-2 continue to use the pre-trained Q-former in conjunction with the \texttt{eva-vit-g} model, thereby narrowing the scope of ViT exploration. In contrast, $\text{EVL}_{\text{Gen}}$ significantly trims the training time while maintaining satisfactory performances, thus providing researchers with the latitude to explore a wider array of advanced ViTs in future investigations.

Furthermore, MACs (FLOPs) in Q-Former and TomeFormer is discussed in Section~\ref{section:macs}.

\begin{table}[!t]
	\centering
        \small
        \begin{tabular}{l l l  c c c c}
        \toprule
        & LLM & Model & C & B4 & M & R \\
        \midrule
        \multirow{4}{*}{\rotatebox[origin=c]{90}{NoCaps}}  & \multirow{2}{*}{OPT} & BLIP-2 & 112.2 & 44.4 & 29.5 & 59.7  \\
        & & $\text{EVL}_{\text{Gen}}$ & \textbf{117.4} & \textbf{45.9} & \textbf{30.3} & \textbf{61.1} \\
        \cmidrule{2-7}
        & \multirow{2}{*}{Vicuna} & BLIP-2 & 115.6 & 45.3 & 30.3 & 60.6  \\
        & & $\text{EVL}_{\text{Gen}}$ & \textbf{119.0} & \textbf{45.9} & \textbf{30.6} & \textbf{61.5}  \\
        \midrule
        \multirow{4}{*}{\rotatebox[origin=c]{90}{Flickr30K}}  & \multirow{2}{*}{OPT} & BLIP-2 & 77.1 & 28.7 & 23.9 & 51.6 \\
        & & $\text{EVL}_{\text{Gen}}$ & \textbf{82.0} & \textbf{30.0} & \textbf{24.5} & \textbf{52.4} \\
        \cmidrule{2-7}
        & \multirow{2}{*}{Vicuna} & BLIP-2 & 80.0 & 30.1 & \textbf{24.8} & 52.1 \\
        & & $\text{EVL}_{\text{Gen}}$ & \textbf{81.8} & \textbf{30.3} & 24.5 & \textbf{52.2} \\
        \bottomrule	
        \end{tabular}
        \caption{Comparison of different models' performance on zero-shot NoCaps and Flickr30K captioning. C$\rightarrow$CIDEr, B4$\rightarrow$BLEU-4, M$\rightarrow$METEOR, R$\rightarrow$ROUGE}
    \label{table:image_caption}
\end{table}

\subsection{Evaluation of $\text{EVL}_{\text{Gen}}$-Video}
We proceed to evaluate the performance of fine-tuned $\text{EVL}_{\text{Gen}}$-Video models in video captioning tasks, utilizing OPT-2.7b as the language model decoder. Our investigation includes two specific variants of $\text{EVL}_{\text{Gen}}$-Video: the first is exclusively pre-trained on image data, while the second is further \textit{enhanced by pre-training on a corpus of 2 million video-text pairs sourced from the WebVid~\citep{bain2021frozen} dataset.} To provide a comprehensive evaluation, we benchmark $\text{EVL}_{\text{Gen}}$-Video against five distinct models, described as follows:
\begin{itemize}[leftmargin=*,noitemsep,nolistsep]
    \item \textbf{Baseline (concat)}: This model processes each frame of a video individually and concatenates their visual features to generate a single prompt for the LLM. This method is analogous to the strategy employed in InstructBLIP.
    \item \textbf{Baseline (mean)}: Similar to the concat baseline, this model processes each video frame individually but averages the visual features to create a single prompt for the LLM.
    \item \textbf{Video-LLaMA}: This variant incorporates the BLIP-2 framework and enhances it with an additional temporal Q-former layer. For this evaluation, we focus solely on the vision-language component of Video-LLaMA.
    \item \textbf{VideoChat}: This model extends BLIP-2 by integrating additional Uniformer modules within the ViT architecture and also incorporates learnable temporal queries in its Q-former component.
    \item \textbf{VideoCoCa}: In this model, we adapt the OpenCoCa framework by mlfoundations and augment the existing CoCa architecture with a learnable attentional pooler, resulting in VideoCoCa.
\end{itemize}

\begin{table}[!t]
    \centering
    \small
        \begin{tabular}{l c c c c}
        \toprule
        \multirow{2}{*}{Models} &  \multirow{2}{*}{C} &  \multirow{2}{*}{B4} &  \multirow{2}{*}{M} &  \multirow{2}{*}{R}\\
        & & & & \\
        \midrule        
        Baseline (concat) & 65.5 & 44.4 & 31.9 & 64.1\\
        Baseline (mean) & 67.8 & 47.3 & 32.2 & 65.0  \\
        $\text{EVL}_{\text{Gen}}$-image & 68.4 & 47.6 & 32.4 & 65.3  \\
        $\text{EVL}_{\text{Gen}}$-video & 69.8 & 48.3 & 32.6 & 65.8  \\
        $\text{EVL}_{\text{Gen}}$-video-scst & \textbf{74.0} & \textbf{49.2} & \textbf{33.0} & \textbf{66.5}  \\
        \midrule
        Video-LLaMA & 59.3 & 47.7 & 29.6 & 63.7  \\
        VideoChat & 58.0 & 46.5 & 29.5 & 63.4  \\
        VideoCoCa (open) & 63.0 & 48.5 & 31.4 & 64.8  \\
        \bottomrule	
        \end{tabular}
        \caption{Comparison of different models' performance on MSR-VTT video captioning. Models are pre-trained using 2 million video-text pairs from WebVid dataset, except for image pre-trained $\text{EVL}_{\text{Gen}}$-image. }
    \label{table:msrvtt}
\end{table}

\paragraph{Evaluation on MSR-VTT}
As detailed in Table~\ref{table:msrvtt}, $\text{EVL}_{\text{Gen}}$-Video demonstrates superior performance relative to the baseline models, even without the aid of video-text pre-training. This result highlights the effectiveness of our proposed \textit{Temporal Attentive Soft Token Contextualizing} in capturing temporal dynamics. Additionally, we observe an enhancement in performance when incorporating video-text pre-training along with Self-Critical Sequence Training (SCST) ~\citep{rennie2017self}.

\textit{Temporal Attentive Soft Token Contextualizing} has the distinct advantage of maintaining the integration of the well-pretrained image-text model (i.e., $\text{EVL}_{\text{Gen}}$-Image). This contrasts with models such as Video-LLaMA and VideoChat, where the original BLIP-2 architecture is altered, necessitating a complex re-alignment process using video-text pairs. Our empirical analysis indicates that such re-alignment is a non-trivial endeavor (as shown in Table~\ref{table:msrvtt}, Video-LLaMA and VideoChat struggle to re-align with 2M WebVid video-text pairs). It is worth noting that our VideoCoCa model is at a disadvantage when benchmarked against Google's reported results, which benefit from extensive training on a much larger billion-scale dataset.

\paragraph{Evaluation on MSVD}
Similarly, we evaluate $\text{EVL}_{\text{Gen}}$'s performance against Video-LLaMA, VideoChat, and VideoCoCa using the MSVD caption dataset (presented in Table~\ref{table:msvd}). Our results corroborate that $\text{EVL}_{\text{Gen}}$ consistently surpasses these competing models, further attesting to its robust performance across different video caption tasks.

\begin{table}[!t]
	\centering
        \small
        \begin{tabular}{l  c c c c}
        \toprule
        Models & C & B4 & M & R \\
        \midrule
        Video-LLaMA  & 121.2 & 61.6 & 40.3 & 77.8  \\
        VideoChat    & 118.4 & 64.1 & 41.0 & 78.7  \\
        VideoCoCa (open)   & 150.9 & 67.7 & 45.3 & 81.9  \\
        $\text{EVL}_{\text{Gen}}$-video  & \textbf{158.2} & \textbf{68.4} & \textbf{46.8} & \textbf{83.1}  \\
        \bottomrule	
        \end{tabular}
        \caption{Comparison of different models' performance on MSVD video captioning.}
    \label{table:msvd}
\end{table}

\section{Ablations and Analysis}\label{sec:analysis}
\subsection{Ablations on TomeFormer}
Within the TomeFormer, the vision-to-language connector in $\text{EVL}_{\text{Gen}}$, we introduce a hyper-parameter $r$ that regulates the number of spatial tokens merged at each layer. Increasing $r$ substantially reduces the token count, but runs the risk of eliminating important visual details. On the other hand, a smaller $r$ produces two main effects: (1) a more diffuse representation of visual features, complicating the optimization landscape, and (2) elongated soft prompts for the LLM, leading to increased computational cost during training, such as memory overflow and extended training duration.

\begin{figure}[!t]
\centering
\includegraphics[width=0.48\textwidth]{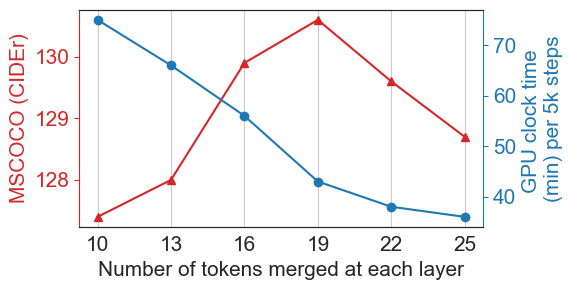}
\caption{Trade-off between MSCOCO captioning scores (depicted in red) and GPU training time (depicted in blue) as a function of the number of tokens merged ($r$) in TomeFormer. }
\label{fig:simvlg_ablation}
\end{figure}

To study the effects of $r$, we conduct an ablation experiment using $8 \times$ RTX-A6000 and the CCS-14M dataset for pre-training. The models are trained for 60,000 steps, and their performance is evaluated using CIDEr scores on MSCOCO captioning. In Figure~\ref{fig:simvlg_ablation}, we observe that a smaller $r$ (e.g., 10) places a higher computational load on both TomeFormer and the LLM, extending training time and compromising optimization, as evidenced by lower CIDEr scores. In contrast, a larger $r$ value (e.g., 25) expedites training but at the expense of model performance, likely due to excessive feature compression and consequent information loss. Additional ablation results on VQAv2, GQA and OKVQA are provided in Appendix~\ref{appendix:ablation_tomeformer}.

\begin{figure}[!t]
\centering
\includegraphics[width=0.48\textwidth]{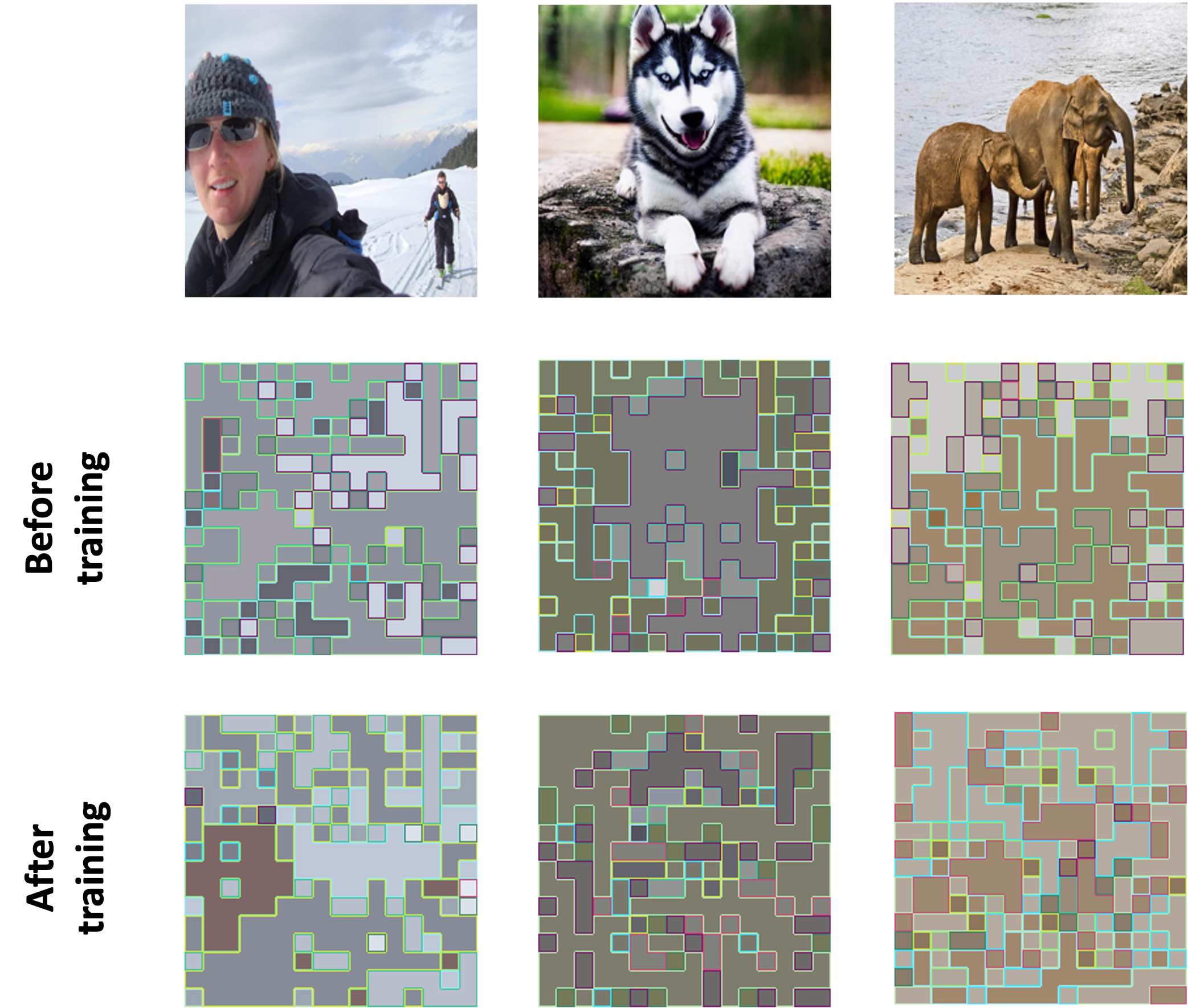}
\caption{Pre- and post-training visualization of merged tokens in $\text{EVL}_{\text{Gen}}$. The visual features compressed via token merging exhibit semantic informativeness even prior to training. This inherent characteristic facilitates $\text{EVL}_{\text{Gen}}$'s ability to converge quickly in an end-to-end training setup.}
\label{fig:simvlg_demo}
\end{figure}

\subsection{Ablations on ViT} Experimental results on $\text{EVL}_{\text{Gen}}$ with different visual encoders (ViT) are provided in Appendix~\ref{appendix:ablation_vit}. $\text{EVL}_{\text{Gen}}$ is robust to different visual encoders, and the stronger ViT generally leads to better results. This implies that while $\text{EVL}_{\text{Gen}}$ also requires retraining for different ViTs, but the single-stage training and quick convergence allow it to benefit from a future release of the latest ViTs, given its capability of fast adaptation.

\subsection{Vicuna-7b as the LLM}
In Table~\ref{table:vicuna}, we provide experimental results of BLIP-2 and ours using Vicuna-7b as the LLM decoder, on zero-shot VQAv2, GQA, OKVQA and MSCOCO captioning (CIDEr) tasks without additional fine-tuning.

$\text{EVL}_{\text{Gen}}$ achieves better performance in GQA, OKVQA, and MSCOCO captioning using considerably less computing, though our model under-performs in VQAv2. As we discussed in Limitations, the inability of $\text{EVL}_{\text{Gen}}$ to extract question-conditioned visual features may lead to inferior results on VQAv2.

\begin{table}[!t]
	\centering
        \scriptsize
        \begin{tabular}{l l l c c c c}
        \toprule
        Models & Data & steps & VQA & GQA & OK & COCO \\
        \midrule
        BLIP-2 & 104M & 330k & \textbf{57.8} & 35.7 & 27.8 & 138.0\\
        $\text{EVL}_{\text{Gen}}$ & 104M &  90k & 53.4 & 34.7 & 30.6 & 137.8\\
        $\text{EVL}_{\text{Gen}}$ & 11M  & 150k & 54.6 & 34.0 & 27.3 & 138.0\\	
        $\text{EVL}_{\text{Gen}}$ & 104M & 150k & 55.5 & \textbf{36.3} & \textbf{30.6} & 137.9 \\
        $\text{EVL}_{\text{Gen}}$ & 104M & 250k & 54.8 & 35.6 & 30.4 & \textbf{139.1}\\
        \bottomrule	
        \end{tabular}
        \caption{Comparison of different models' performance on zero-shot VQA and MSCOCO captioning (CIDEr) tasks without additional fine-tuning. Both BLIP-2 and E2VLGen use Vicuna-7b as the LLM decoder.}
    \label{table:vicuna}
\end{table}

\subsection{Token Merging Visualization in $\text{EVL}_{\text{Gen}}$}
One notable advantage of $\text{EVL}_{\text{Gen}}$ over BLIP-2 is the absence of a requisite Stage-1 pre-training for the vision-to-language connector. This simplifies the training pipeline by removing the need to train the model to extract text-informative visual features. We posit that the token merging process in TomeFormer naturally aggregates tokens associated with visually similar elements, thereby yielding concise yet semantically rich visual features from the onset of training. This inherent capability allows $\text{EVL}_{\text{Gen}}$ to benefit from a more streamlined, single-stage training regimen with just one learning objective.

Essentially, our token merging strategy serves as an efficient approximation of Q-Former's functionality, compressing visual features in a semantically meaningful manner. Figure~\ref{fig:simvlg_demo} illustrates this, displaying the visual tokens before and after training with our TomeFormer. The figure shows that the compressed visual features obtained via token merging are semantically informative and offer basic object segmentation within the image. Furthermore, the semantic coherence of these merged tokens improves as training advances. Additional visualization examples are shown in Appendix~\ref{appendix:additional_demo}.

\subsection{MACs (FLOPs) in Q-Former and TomeFormer}\label{section:macs}
In this section, we compute \textbf{multiply–accumulate operations} (MACs) in Q-Former and TomeFormer. MACs performs $a \leftarrow a + (b \times c)$. Whereas, FLOPs is \textbf{floating operations} which includes $\times$ / $+$ / $\div$ ... etc. One MACs has one $\times$ and one $+$. And thus, roughly speaking, FLOPs is two times as MACs.

In our experiments, BLIP-2 and $\text{EVL}_{\text{Gen}}$ have identical ViTs and LM decoders. Thus, we only compare the MACs in VL Connector in BLIP-2 and $\text{EVL}_{\text{Gen}}$ (i.e., Q-Former and TomeFormer). 

There's a large MACs in BLIP-2 stage-1 due to three forward passes using Q-Former, where the last forward-pass used for caption loss dominates (27.0G). In contrast, $\text{EVL}_{\text{Gen}}$ does not require such a representation training stage (stage-1) at all.

Another reason why BLIP-2 stage-1 is slow is that the computation of Image-Text Contrasive and Image-Text Matching losses needs \texttt{concat\_all\_gather} operations that require GPU communications. Further Image-Text Matching requires binomial sampling of hard negatives. In comparison, our $\text{EVL}_{\text{Gen}}$ circumvents such computations/communications.

\begin{table}[!t]
    \centering
    \small
        \begin{tabular}{l c r c r }
        \toprule
        \multirow{2}{*}{Models} & \multirow{2}{*}{\makecell[c]{Stage 1 \\ (MACs)}} & \multirow{2}{*}{\makecell[c]{Stage 1 \\steps}} & \multirow{2}{*}{\makecell[c]{Stage 2 \\ (MACs)}} & \multirow{2}{*}{\makecell[c]{Stage 2 \\steps}} \\
        \\
        \midrule
        BLIP-2  &  36.7G & 250k & 6.28G &  80k \\
        $\text{EVL}_{\text{Gen}}$ & - & -  & 11.9G & 250k     \\
        $\text{EVL}_{\text{Gen}}$ & - & -  & 11.9G & 150k     \\
        $\text{EVL}_{\text{Gen}}$$_{\text{55M}}$ & - & -  & 5.6G & 90k  \\
        \bottomrule	
        \end{tabular}
        \caption{\textbf{Multiply–accumulate operations} (MACs) comparison of Q-Former (of BLIP-2) and TomeFormer (of $\text{EVL}_{\text{Gen}}$) when utilizing OPT-2.7b as the LLM. }
    \label{table:macs}
\end{table}

\section{Discussion and Conclusion}
This paper introduces $\text{EVL}_{\text{Gen}}$, an efficient and streamlined pre-training framework for vision-language generative models. Like BLIP-2, $\text{EVL}_{\text{Gen}}$ employs frozen ViT and LLM. It further leverages a conventional Transformer architecture with token-merging capabilities, known as TomeFormer, to act as the vision-to-language connector. Compared to BLIP-2, $\text{EVL}_{\text{Gen}}$ offers the distinct advantage of one-stage training. This reduces computational overhead and maintains competitive performance even with only $1/3$ to $1/6$ of the computational budget required by BLIP-2.

We have also extended $\text{EVL}_{\text{Gen}}$'s applicability to video captioning tasks by incorporating the \textit{Temporal Attentive Soft Token Contextualizing} into its ViT. This enhances the model's temporal modeling capabilities, culminating in the creation of $\text{EVL}_{\text{Gen}}$-Video. This extension has proven efficacious, delivering commendable performance even without specialized video-text pre-training. Our investigation underscores that a temporal module, which does not disrupt the integration of the well-pretrained image-text model (e.g., BLIP-2 and $\text{EVL}_{\text{Gen}}$), is a key factor contributing to this success.

$\text{EVL}_{\text{Gen}}$ demonstrates the possibility of achieving state-of-the-art performance in vision-language tasks without the need for complex training regimens or high computational budgets. This work thus makes a significant contribution to the ongoing efforts to develop more accessible, efficient, and powerful models for understanding and generating visual and textual information.

\section*{Limitations}
While $\text{E}{2}\text{VL}{\text{Gen}}$ has showcased its capacity for rapid convergence in VLM pre-training and has demonstrated notable proficiency in zero-shot image/video captioning, certain limitations warrant consideration.
\begin{itemize}[leftmargin=*,noitemsep,nolistsep]
    \item Our approach is guided by a straightforward design aimed at facilitating the efficient and effective training of VLMs. To maintain simplicity in our methodology, we adopt a fixed value of $r$ (19) within TomeFormer to compress visual information (i.e., a fixed length of visual soft-prompt). However, it is worth acknowledging that various images or videos might benefit from distinct optimal compression rates ($r$). Consequently, the incorporation of a variable $r$ (i.e., variable length of soft-prompts for language models) may be deemed more desirable (A similar concern is present in BLIP-2, where the length of soft-prompts is consistently set to 32.).
    \item One trade-off associated with the simplistic design of TomeFormer is its inability to enable text-specific selection of visual features. In applications like VQA, extracting visual features conditioned on the accompanying questions is considered beneficial. However, the current configuration of TomeFormer lacks the provision for this text-conditioned property within VLMs. A prospective redesign of TomeFormer that incorporates text-conditioned visual feature selection holds the potential to enhance VQA performance.
\end{itemize}

\section*{Ethics Statement}
This research aims to enhance both the efficiency and applicability of vision-language generative models via $\text{EVL}_{\text{Gen}}$. Although our research does not involve human subjects directly, it is important to acknowledge and discuss the broader ethical implications.
\paragraph{Data Bias and Fairness:} Our model is trained on publicly available datasets, namely CapFilt, MSCOCO, MSR-VTT, MSVD, and WebVid. While these datasets are widely used, we acknowledge that we cannot fully ascertain the extent to which they may contain discriminatory, biased, or sensitive material. Given that our model inherits the biases present in these training datasets, there exists the risk of perpetuating or even amplifying existing societal biases. Despite the broad acceptance of these datasets, caution should be exercised.
\paragraph{Real-world Deployment and Responsible Usage:} Like all generative models, $\text{EVL}_{\text{Gen}}$ could be misappropriated for creating misleading or harmful content. Thus, it is imperative to implement safety mechanisms to counter such misuse when deploying the model in real-world applications. Special attention should also be paid to ensure that the model does not inadvertently produce outputs that could disclose sensitive or personal information.
Finally, while $\text{EVL}_{\text{Gen}}$ is intended as a general-purpose model, its application in contexts that could worsen societal biases or spread misinformation is a pressing concern. Developers and researchers employing $\text{EVL}_{\text{Gen}}$ are advised to be cognizant of these risks and consider incorporating fairness-aware or truth-aware components into their systems.

\bibliography{acl_latex}

\appendix

\section{Additional Related Works}\label{appendix:additional_ref}
\paragraph{Image-Language Models} Vision-language models generally fall into two categories: dual-encoder models and fusion-encoder models. Pioneering works like CLIP~\citep{radford2021learning} and ALIGN~\citep{jia2021scaling} serve as exemplary dual-encoder models, demonstrating exceptional performance in zero-shot classification tasks. These architectures also excel in image-text retrieval, as their features can be pre-computed and stored, allowing for \textit{efficient similarity score computation via dot-product operations.} Fusion-encoder models~\citep{lu2019vilbert, tan2019lxmert, alayrac2022flamingo, dou2022empirical, li2022blip, dou2022coarsetofine, xu2022bridge}, such as ALBEF~\citep{li2021align}, mPLUG~\citep{li-etal-2022-mplug}, X-VLM~\cite{zeng2022multi}, and VLMo~\citep{bao2022vlmo}, employ cross-attention mechanisms to enable deep interactions between visual and linguistic features. Other designs include concatenating features of each modality before feeding them into a Transformer~\citep{chen2020uniter, li2020oscar, zhang2021vinvl, gan2020large, li2021unimo, cho2021unifying, huang2020pixel, huang2021seeing, shen2022how, kamath2021mdetr, yang2022unitab, wang2022ofa, kim2021vilt, xue2021probing, wang2022git, wang2022image, xu-etal-2021-e2e}. These models excel in complex tasks like closed-set Visual Question Answering (VQA) and visual entailment.

BLIP-2~\cite{li2023blip} is proposed to leverage pre-trained frozen ViTs and LLMs to alleviate the computation demands in the full end-to-end training. Under this learning paradigm, \citet{zhang2023vpgtrans} introduce visual-prompt transfer learning to mitigate visual-language re-alignment cost when using different LLMs. \citet{jian2023bootstrapping} propose decoupled language pre-training to alleviate the intensive data requirement in BLIP-2.

More recently, VLM research \cite{yin2023survey, fu2023mme} also explores visual instruction tuning \cite{xu-etal-2023-multiinstruct, liu2023visual, li-etal-2022-mplug, dai2023instructblip}, multi-modal in-context learning \cite{chen2022learning, he2023icl, shao2023prompting, gupta2023visual, yang2022empirical} and Chain-of-Thought prompting \cite{Zheng_NeurIPS2023, lu2022learn}, interleaved image-text generation \cite{aiello2024jointly}, and hallucination \cite{li2023evaluating}.

\paragraph{Visual Redundancy}
The concept of visual redundancy plays a pivotal role in the field of computer vision. It pertains to the phenomenon where semantic information is conveyed through multiple channels, often involving the use of various visual elements like shape and color to represent complex symbols. Recognizing the impact of this redundancy on deep learning algorithms, there has been a shift towards devising methods to minimize it, thereby enhancing efficiency. For example, IA-RED2~\cite{NEURIPS2021_d072677d} has an interpretable design to dynamically and gracefully remove redundant tokens. In \citet{9878548}, the redundancy of ViT is discussed on embedding, attention, and weight levels. Eventful Transformer~\cite{Dutson_2023_ICCV} discusses the temporal redundancy in the temporal dimension.

\section{BLIP-2 Training Configurations}\label{appendix:blip2-training-details}
We re-train BLIP-2 from scratch using LAION-115M, CCS-14M, MSCOCO from CapFilt dataset~\cite{li2022blip}. The models are trained on eight A100-80G GPUs. The training configurations are shown in Table~\ref{table:blip2-configs}.

\begin{table}[!t]
	\centering
        \small
        \begin{tabular}{l c c c c c}
        \toprule
        Phase & Bsz & init LR & min LR & steps & warm-up \\
        \midrule
        Stage-1 & 1600 & $1e^{-4}$ & $1e^{-5}$ & 250k & 5k \\
        Stage-2 & 1600 & $1e^{-4}$ & $1e^{-5}$ & 80k & 5k \\
        \bottomrule	
        \end{tabular}
        \caption{Training configurations for our re-produced BLIP-2.}
    \label{table:blip2-configs}
\end{table}

\section{Technical Details of Token Merging}\label{appendix:tome_details}
In this section, we briefly summarize the technical designs of Token Merging (ToMe)~\cite{bolya2022tome}. Token Merging was initially proposed in \citet{bolya2022tome} for accelerating ViTs without training. Whereas we re-purpose ToMe to condense the visual features used as language prompts in the LLM. Please refer to Sections 3 of \citet{bolya2022tome} for full details.

\paragraph{Strategy.} The token merging operations take place in between the attention and MLP blocks of each Transformer layer. ToMe reduces $r$ tokens per layer. And over the $L$ layers of a Transformer, it reduces a total of $r \times L$ tokens. In our experiments, we set $r=19$ and our TomeFormer has 12 layers.

\paragraph{Token Similarity.} The similarities of tokens are defined by the cosine similarity (dot product) of keys of tokens.

\paragraph{Bipartite Matching.} The bipartite soft matching algorithm is summarized as follows:
\begin{itemize}[leftmargin=*,noitemsep,nolistsep]
    \item Tokens are randomly partitioned into two sets $\mathbb{A}$ and $\mathbb{B}$.
    \item Each token in set $\mathbb{A}$ is linked to the most similar token in set $\mathbb{B}$.
    \item Keep links with top $r$ similarities.
    \item Merge tokens with top $r$ links. 
    \item Concatenate set $\mathbb{A}$ and $\mathbb{B}$ back into a single set.
\end{itemize}

\section{Details on Our Implementations of BLIP-2 and VideoCoCa}
Our reported results of our re-trained BLIP-2 are slightly worse than what was reported in \citet{li2023blip}. There are mainly three reasons:

\begin{itemize}[leftmargin=*,noitemsep,nolistsep]
    \item We are only able to download 104M image-text pairs from the original 129M CapFlit dataset.
    \item We intentionally exclude the VG dataset from our pre-training procedure, as it mainly consists of localized captions. Thus, our re-trained BLIP-2 is more challenging when evaluated on GQA, which is built on VG dataset.
    \item The exact dataset weighting is unknown from the LAVIS project, we use a weighting that is based on the size of each pre-training dataset, i.e., CSS14M, LAION115M, MSCOCO.
\end{itemize}

For video captioning in Table~\ref{table:msrvtt} and Table~\ref{table:msvd}, because VideoCoCa is not open-sourced, we use a pre-trained model \texttt{OpenCoCa} released by mlfoundations.

\begin{figure}[!t]
\centering
\includegraphics[width=0.48\textwidth]{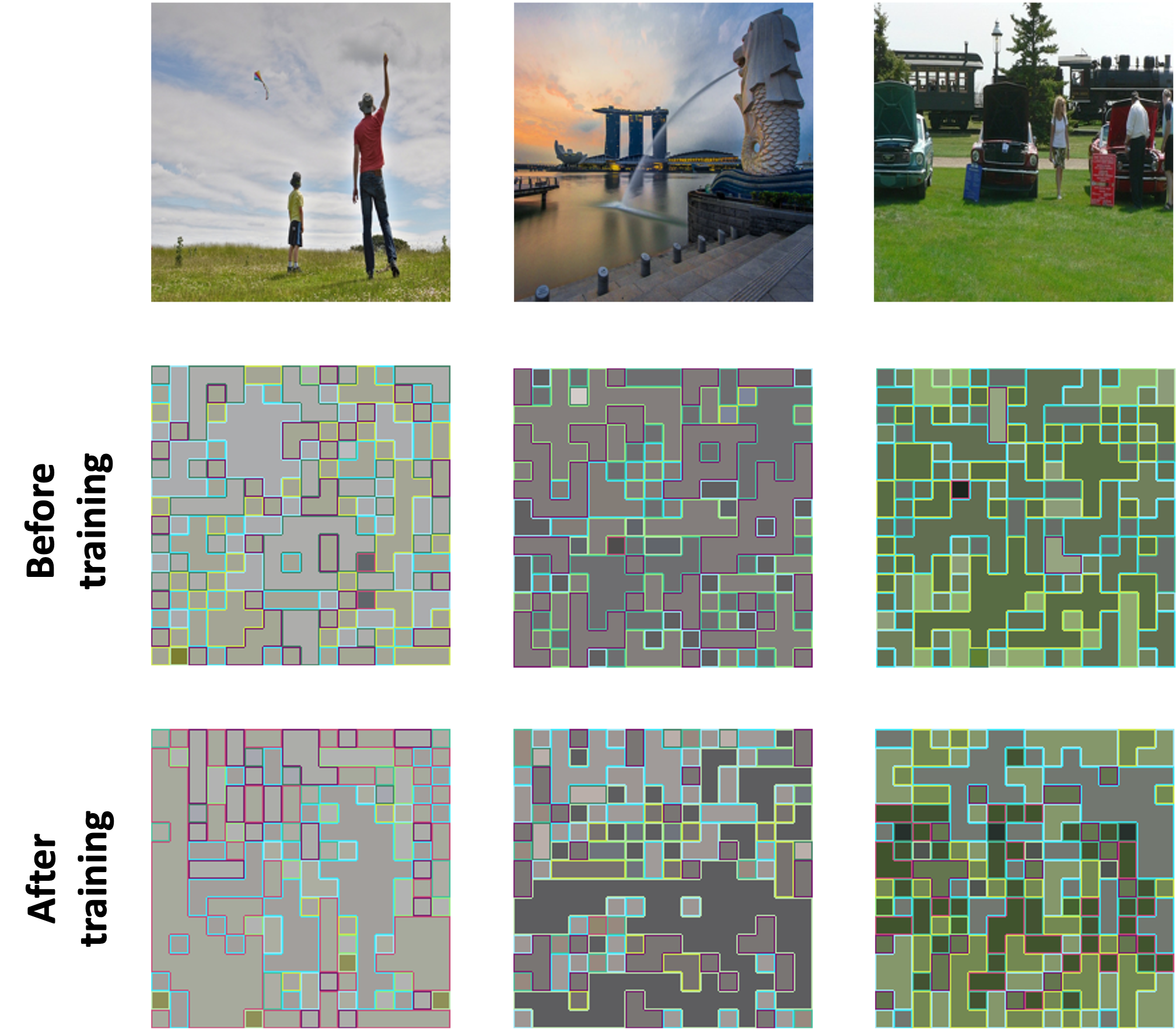}
\caption{Additional pre- and post-training visualization of merged tokens in $\text{EVL}_{\text{Gen}}$.}
\label{fig:simvlg_demo2}
\end{figure}

\section{Additional Token Merging Visualization in $\text{EVL}_{\text{Gen}}$}\label{appendix:additional_demo}

In this section, we provide additional examples of token merging visualization (before and after training) in $\text{EVL}_{\text{Gen}}$ in Figure~\ref{fig:simvlg_demo2}.

The visual features compressed via token merging in the TomeFormer exhibit semantic informativeness even prior to training. This inherent characteristic facilitates $\text{EVL}_{\text{Gen}}$'s ability to converge quickly in an end-to-end training setup.

\section{Ablation on Visual Encoders}\label{appendix:ablation_vit}

One of the limitations of BLIP-2 is that it requires an extensive stage-1 pre-training for every different vision encoder. This prohibits practitioners from exploring stronger ViTs when they are available. $\text{EVL}_{\text{Gen}}$ offers fast training of models, allowing for exploration of different ViTs as visual encoders.

We conduct an ablation experiment on two ViTs (CLIP$_\text{L}$ and EVA-ViT$_\text{G}$) using $8 \times$ RTX-A6000 and the CCS-14M dataset for pre-training. The models are trained for 60,000 steps. 

Shown in Table~\ref{table:ablation_vit}, $\text{EVL}_{\text{Gen}}$ is robust to different visual encoders, and the stronger ViT leads to better results. This implies that while $\text{EVL}_{\text{Gen}}$ also requires retraining for different ViTs, but the single-stage training and quick convergence allow it to benefit from a future release of the latest ViTs, given its capability of fast adaptation.

\begin{table}[!t]
	\centering
        \small
        \begin{tabular}{l l c c c c}
        \toprule
        LLM & ViT & VQA & GQA & OK & COCO \\
        \midrule
        \multirow{2}{*}{OPT} & CLIP$_\text{L}$ & 44.7 & \textbf{30.9} & 22.7 & 123.9  \\
        & EVA-ViT$_\text{G}$ & \textbf{45.2} & 30.6 & \textbf{22.8} & \textbf{130.6} \\
        \midrule
        \multirow{2}{*}{Vicuna} & CLIP$_\text{L}$ & 49.0 & 33.0 & 23.6 & 125.2\\
        & EVA-ViT$_\text{G}$ & \textbf{52.5} & \textbf{34.6} & \textbf{27.9} & \textbf{132.4} \\
        \bottomrule	
        \end{tabular}
        \caption{Ablation studies on different visual encoders of $\text{EVL}_{\text{Gen}}$. VQA$\rightarrow$VQAv2, OK$\rightarrow$OKVQA, COCO$\rightarrow$MSCOCO (CIDEr).}
    \label{table:ablation_vit}
\end{table}

\begin{table}[!t]
	\centering
        
        \begin{tabular}{l c c c c}
        \toprule
        $r$ & VQA & GQA & OK & \textbf{COCO} \\
        \midrule
        10 & 45.7 & 31.3 & 23.6 & 127.5  \\
        13 & 46.2 & 31.4 & 24.5 & 128.0 \\
        16 & 46.3 & 30.9 & 24.3 & 129.9 \\
        19 & 45.2 & 30.7 & 22.8 & 130.6 \\
        22 & 45.5 & 31.5 & 21.8 & 129.7 \\
        25 & 44.7 & 31.1 & 21.5 & 128.7 \\
        \bottomrule	
        \end{tabular}
        \caption{Ablation studies on $r$ in TomeFormer.}
    \label{table:ablation_tomeformer}
\end{table}

\section{Ablations on TomeFormer}\label{appendix:ablation_tomeformer}
In this section, we provide experimental results in VQAv2, GQA, and OKVQA of $\text{EVL}_{\text{Gen}}$, by varying hyper-parameter $r$ in TomeFormer. As we can see from Table~\ref{table:ablation_tomeformer}, $\text{EVL}_{\text{Gen}}$ is robust to the choice of $r$.

\section{The Use of the AI Assistants}
We use ChatGPT for grammar correction and a bit of sentence-level polishing of some of our writing.

\end{document}